\pdfoutput=1
\documentclass[twocolumn]{ceurart}

\usepackage{anyfontsize}
\usepackage{listings}
\usepackage{booktabs}
\usepackage{hhline}
\usepackage{tabularx}
\usepackage{graphicx}
\usepackage{diagbox}
\usepackage{indentfirst}
\usepackage{subcaption}
\usepackage{float}
\usepackage{placeins}
\RequirePackage{etoolbox}

\newcolumntype{?}{!{\vrule width 1pt}}
\begin{document}
\fontsize{9}{10}\selectfont 

\copyrightyear{2024}
\copyrightclause{}

\conference{}

\title{YOWOv3: An Efficient and Generalized Framework for Human Action Detection and Recognition}

\tnotemark[1]
\tnotetext[1]{}

\author[1]{Nguyen Dang Duc Manh}[%
orcid=0009-0002-0717-5439,
email=22520847@gm.uit.edu.vn,
url=https://github.com/Hope1337,
]
\author[1]{Duong Viet Hang}[%
orcid=0000-0002-9728-4438,
email=hangdv@uit.edu.vn ,
]
\author[2]{Jia Ching Wang}[%
orcid=0000-0003-0024-6732,
email=jcw@csie.ncu.edu.tw,
]
\author[2]{Bui Duc Nhan}[%
orcid=0009-0001-0204-8142,
email=nhanbuiduc.work@gmail.com,
]
\address[1]{University of Information Technology, Ho Chi Minh City, Viet Nam}
\address[2]{National Central University, Taoyuan City, Taiwan}

\newcommand{\commiturl}{\url{https://github.com/AakiraOtok/YOWOv3}}
\begin{abstract} 
	\justifying{Our new framework, YOWOv3, is an enhanced version of YOWOv2 that we provide in this study with a focus on Spatio Temporal Action Detection task. This framework is made by offering a more accessible approach to experiment deeply with different configurations and to easily customize different model components, which minimizes the amount of labor needed to comprehend and alter the source code. YOWOv3 outperforms YOWOv2 on two popular datasets (UCF101-24 and AVAv2.2) for Human Action Detection and Recognition. In particular, the prior model, YOWOv2, with 109.7M parameters and 53.6 GFLOPs, obtains a mAP of 85.2\% and 20.3\% on UCF101-24 and AVAv2.2, respectively. On the other hand, our model, YOWOv3, obtains a mAP of 20.31\% on AVAv2.2 and 88.33\% on UCF101-24, by utilizing just 39.8 GFLOPs and 59.8M parameters. The outcomes show that YOWOv3 achieves equivalent performance with a significant reduction in the number of parameters and GFLOPs.}

The code is publicly available at: \commiturl.
\end{abstract}


\maketitle

\section{Introduction}

Spatio Temporal Action Detection (STAD) is a common task in computer vision that involves detecting the location (bounding box), timing (exact frame), and type (class of action) of activities, necessitating the modeling of both spatial and temporal features. STAD finds extensive applications across various fields, playing a crucial role due to its practical significance. Some notable applications include security surveillance, monitoring and preventing school violence, child abuse detection, domestic violence prevention, supporting healthcare applications, virtual reality, and a myriad of other applications. To address the STAD problem, numerous studies have applied such common approach like using the Vision Transformer (ViT) model \cite{hiera, videomaev2, memvit, mvit, mvitv2}. In terms of detection performance, often measured by mean Average Precision (mAP), ViT models have achieved outstanding mAP scores, leading on benchmark datasets. However, a drawback of ViT models is their requirement for massive computational power. For example, the Hiera model \cite{hiera} has over 600 million parameters, or the VideoMAEv2
\cite{videomaev2} model has up to 1 billion parameters. The enormous parameter count is directly proportional to the GLOPs (Giga Floating Point Operations per second) metric, which can increase to hundreds or even thousands. This leads to significantly increased training overhead (two weeks on 60 A100 GPU for training \cite{videomaev2}) and inference times, requiring powerful processors and limiting the applicability of these models to practical application, where real-time processing capability is always demanded.
% TODO: \usepackage{graphicx} required
\begin{figure}[h]
	\centering
	\includegraphics[width=1.\linewidth]{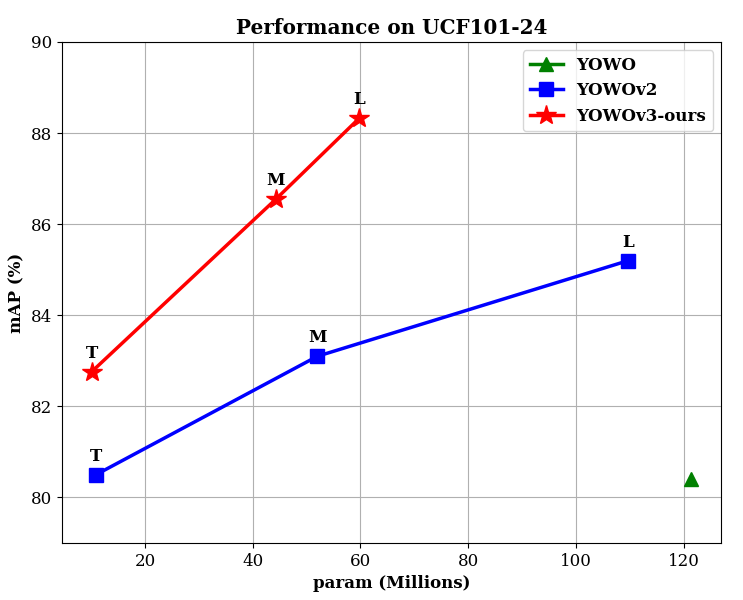}
	\caption{Trade-off between parameters and mAP on UCF101-24. YOWOv3 proves to be an efficient model by enhancing performance while still utilizing computational resources better than previous models.}
	\label{trade-off}
\end{figure}

To tackle the STAD problem while maximizing the mitigation of two drawbacks related to training and inference time, previous studies \cite{yowo} have utilized the Two-Stream Network architecture \cite{twostream} to create a model called YOWO. Moreover, in an effort to advance prior research, an enhanced version of YOWO, referred to as YOWOv2, was introduced \cite{yowov2}. Although both YOWO and YOWOv2 exhibit lower performance on the mAP scale compared to Vision Transformer models, they effectively alleviate the computational demand, fully capable of meeting real-time requirements.

However, both YOWO and YOWOv2 still harbor some lingering limitations: For YOWO, despite being a pioneering lightweight one-stage detector model in the STAD task, its rather simple architecture, coupled with certain techniques, has become outdated and shown to be less effective compared to recently proposed methods. As for YOWOv2, as a successor to its predecessor, it incorporates several new improvements, enhancing the mAP scores on STAD benchmarks. Although the number of parameters has been reduced compared to YOWO, the significantly more complex new architecture has led to a notable increase in GLOPs, contradicting the initial goal of creating an efficient model with low computational requirements. Furthermore, the authors of these two frameworks have discontinued support despite numerous questions raised by the community. This presents significant challenges for future research endeavors. Therefore, the research community is in dire need of a new framework for STAD.

Motivated by the aforementioned reasons, we have developed the new framework called YOWOv3. It demonstrates high efficiency by not only improving performance but also significantly reducing computational resource requirements: reducing the number of parameters by 45.5\% and GLOP by 25.74\% compared to its predecessor. Furthermore, we also provide a plethora of pretrained resources for quick finetuning, catering to practical applications, saving significant training time, and contributing to facilitating research on lightweight models for STAD in the future. In conclusion, our contributions are as follows:

\begin{itemize}
	\item \textbf{New framework for STAD:}  Our proposal introduces a new lightweight framework for STAD - called YOWOv3, aiming to serve practical applications and future research.	
	\item \textbf{Efficiency model: } Our model not only improves performance on benchmarks compared to YOWO and YOWOv2 but also efficiently utilizes computational resources, making it entirely capable of meeting real-time application requirements.
	\item \textbf{Multiple pretrained resources for application:} To streamline the practical application of YOWOv3, we have extensively experimented and analyzed various model configurations, creating a range of pretrained resources spanning from lightweight to sophisticated models to cater to diverse requirements for real-world applications. These pretrained resources serve as valuable starting points, enabling users to quickly bootstrap their projects and adapt YOWOv3 to their specific applications.
\end{itemize}

\section{YOWO AND YOWOV2}

\textbf{YOWO:} is the first and only single-stage architecture that achieves competitive results on AVAv2.2 \cite{yowo}. YOWO emerged as a solution for the STAD problem by leveraging the Two-Stream Network architecture \cite{twostream}. While it exhibited commendable performance on benchmark datasets compared to models of similar scale during its time, the simplicity of the YOWO architecture has become outdated. As a result, it lacks the sophistication and advancements seen in contemporary models, limiting its applicability and performance in current STAD research.\\

\textbf{YOWOv2:} Building upon its predecessor, YOWOv2 was introduced as an enhanced version of the YOWO framework \cite{yowov2}. This iteration incorporates novel techniques such as anchor-free object detection and feature pyramid networks, showcasing improved performance compared to the original YOWO model.However, YOWOv2, on the contrary, increases the computational requirements (GLOP), contradicting its initial purpose of creating an efficient lightweight model. This leading to inefficient utilization of computational resources. Furthermore, the authors no longer support YOWOv2, making it difficult for researchers to use and expand this framework in future STAD studies.
\section{PROPOSED FRAMEWORK}

\subsection{OVERVIEW}
\begin{figure*}[h]
	\centerline{\includegraphics[width=40pc]{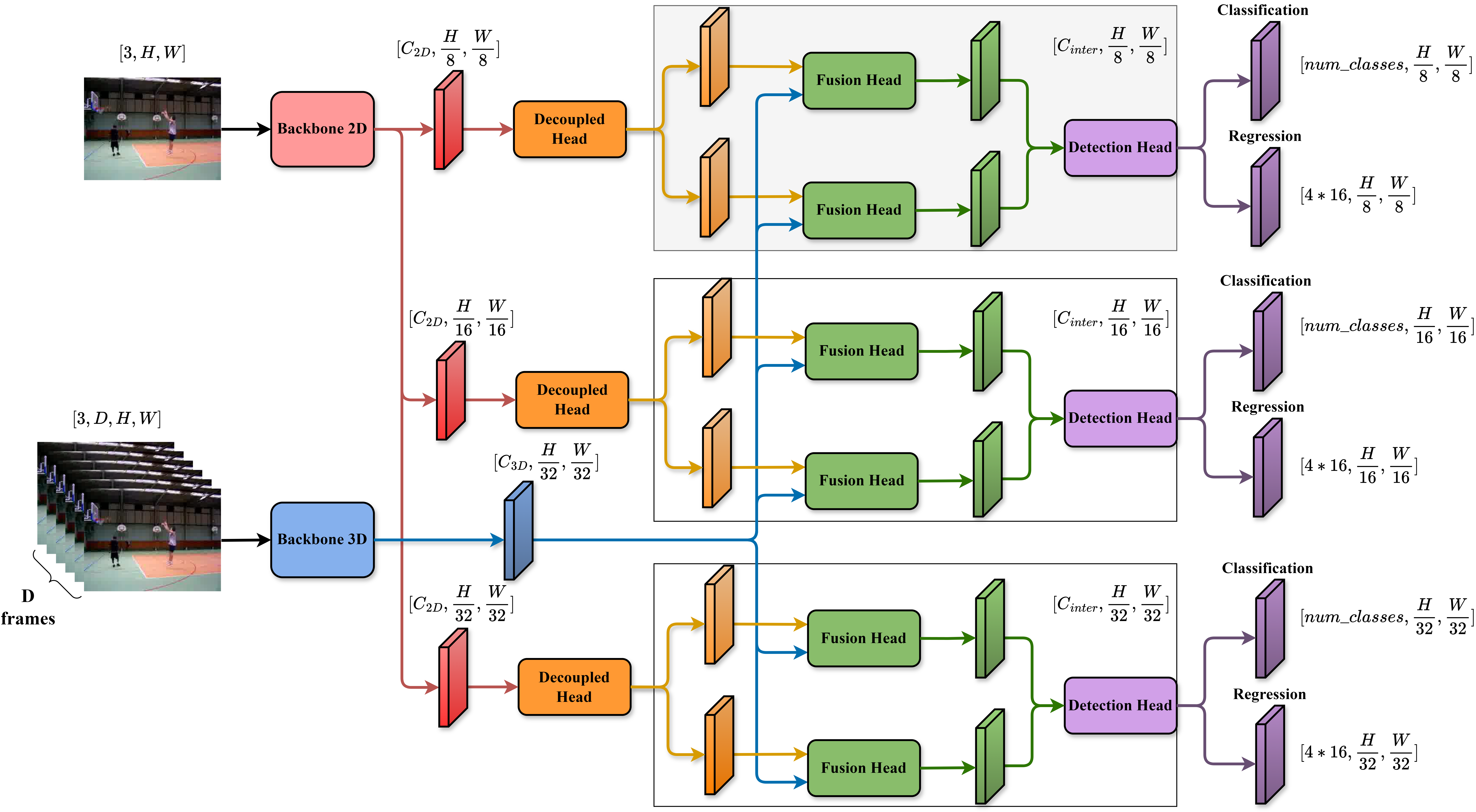}}
	\caption{An overview architecture of YOWOv3}\vspace*{-5pt}
	\label{architect}
\end{figure*}

Figure \ref{architect} presents an overview description of the architecture of YOWOv3. This architecture adopts the concept of the Two-Stream Network, which consists of two processing streams. The first stream is responsible for extracting spatial information and context from the image using a 2D CNN network. The second stream, implemented with a 3D CNN network, focuses on extracting temporal information and motion. The outputs from these two streams are combined to obtain features that capture both spatial and temporal information of the video. Finally, a CNN layer is employed to make predictions based on these extracted features.

Each component in Figure \ref{architect} is assumed to have a distinct function and plays a crucial role in the overall processing flow. We will provide detailed explanations for each module right below.

\subsubsection{Introduction}

We introduce some notations to easily explain and track:

\begin{itemize}
	\item $F: [a, b, c]$ : signifies that tensor F has a shape $[a, b, c]$.
	\item $F_{lv}$ : where $lv \in {lv1, lv2, lv3}$ represent the features from the corresponding levels. 
	\item $H_{lv}$, $W_{lv}$ : represents the height and width at the corresponding level of $F_{lv}$. For example, $F_{lv1}$ will have $H_{lv1} = \frac{H}{8}$ and $W_{lv1} = \frac{W}{8}$, $F_{lv2}$ will have $H_{lv2} = \frac{H}{16}$ and $W_{lv2} = \frac{W}{8}$, and $F_{lv3}$ will have $H_{lv3} = \frac{H}{32}$ and $W_{lv3} = \frac{W}{32}$.
	\item Subscription $cls$, $box$ : object come from classification branch and regression branch respectively. 
	\item $Conv(x)=\text{SiLU}(\text{BatchNorm}(\text{conv2D}(x)))$

\end{itemize}

\subsection{Spatial Feature Extractor} 
As previously mentioned, the model requires a spatial feature extractor to accurately provide information about the locations where actions take place. To fulfill this purpose, we employ the YOLOv8 model \cite{yolov8} - a highly popular convolutional network in the computer vision research community known for its high performance in object detection tasks on reputable and widely recognized benchmarks. Additionally, YOLOv8 boasts a simple yet effective architecture that is easily customizable. We exclude the detection layer at the end while retaining the remainder of the architecture. The input to this module is a feature map with dimensions of $[3, H, W]$, representing the final frame of the input video. By utilizing a pyramid network architecture, the output comprises three feature maps at three distinct levels : $F_{lv1} : [C_{2D}, \frac{H}{8}, \frac{W}{8}]$, $F_{lv2} : [C_{2D}, \frac{H}{16}, \frac{W}{16}]$, $F_{lv3} : [C_{2D}, \frac{H}{32}, \frac{W}{32}]$.
\subsection{Decoupled Head}  

The Decoupled Head is responsible for separating the tasks of classification and regression. The YOLOX model research team discovered that in earlier models, employing a single feature map for both classification and regression tasks made training more challenging \cite{yolox}. Therefore, we have adopted a similar approach to the authors by employing two independent CNN streams for each task to enhance the model's comprehension as outlined below:

\begin{align}
	F_{cls} &= Conv_{cls2}(Conv_{cls1}(x)) \\
	F_{box} &= Conv_{box2}(Conv_{box1}(x))
\end{align}

Remember that the output of the 2D backbone consists of three feature maps at three different levels. Each feature map is fed into the Decoupled Head to generate two feature maps for the classification and regression tasks. The input to the Decoupled Head is a tensor $F_{lv}: [C_{2D}, H_{lv}, W_{lv}]$. The output consists of two tensors with the same shape : $F_{lv} : [C_{inter}, H_{lv}, W_{lv}]$.

\subsection{Temporal Motion Feature Extractor}

To enhance the accuracy of predicting action labels, in addition to relying on contextual and spatial information, we also require a temporal motion information extraction module - referred to as backbone3D - to bolster the model's predictive capabilities for complex action classes. We leverage 3D CNN models provided by the authors Okan Kop¨ukl¨u et al. \cite{3D_CNN_resources}. These 3D CNN models are derived from renowned 2D CNN models and have undergone evaluations on action classification tasks. Additionally, we employ the i3d model \cite{i3d} trained on the same similar task. Input to the 3D backbone is a tensor $F_{3D}:[3, D, H, W]$, which is the whole video, and output is only one tensor $F_{3D}:[C_{3D}, 1, \frac{H}{32}, \frac{W}{32}]$.

\subsection{Fusion Head} 

\begin{figure*}[h]
	\centerline{\includegraphics[width=40pc]{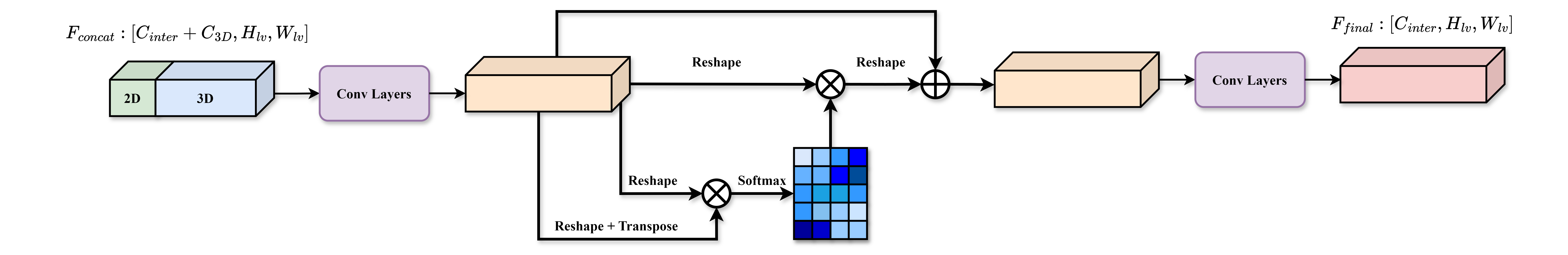}}
	\caption{Overview of Channel Fusion and Attention Mechanism (CFAM) - an attention mechanism in YOWO}\vspace*{-5pt}
	\label{fusion-module}
\end{figure*}

The Fusion Head is responsible for integrating features from both the 2D CNN and 3D CNN streams. The input to this layer consists of two tensors: $F_{lv}: [C_{inter}, H_{lv}, W_{lv}]$ and $F_{3D}: [C_{3D}, 1, \frac{H}{32}, \frac{W}{32}$]. Firstly, $F_{3D}$ is squeezed to obtain a shape of $[C_{3D}, \frac{H}{32}, \frac{W}{32}]$. Then, it is upscaled to match the dimensions of $H_{lv}$ and $W_{lv}$. Next, $F_{3D}$ and $F_{lv}$ are concatenated to obtain the tensor $F_{concat}$: $[C_{inter} + C_{3D}, H_{lv}, W_{lv}]$. Afterward, $F_{concat}$ will be passed into the CFAM module, which is an attention mechanism used in the YOWO model. Figure \ref{fusion-module} provides an overview of the CFAM module. Output of the CFAM module is a feature map $F_{final}:[C_{inter}, H_{lv}, W_{lv}]$

\subsection{Detection Head} 

The Detection Head is responsible for providing predictions for bounding box regression and action label classification. In the study conducted by Xiang Li et al. \cite{generlized_focal_loss}, the authors pointed out that predicting bounding boxes using a Dirac distribution made the model more difficult to train. As a result, the authors proposed allowing the model to learn a more general distribution instead of simply regressing to a single value (Dirac distribution), we adopt the proposed idea. Furthermore, to reduce the model's dependence on selecting hyperparameters for predefined bounding boxes as in previous studies, we also apply the anchor free mechanism \cite{fcos} as in YOLOX \cite{yolox}. 

Input to Detection Head consists of two tensors $F_{cls}$ and $F_{box}$ for classification and regression tasks respectively. The final prediction is generated through a series of convolutional layers as described below:

\begin{align}
	Predict_{cls} &= \text{conv}(Conv_{cls2}(Conv_{cls1}(F_{cls}))) \\
	Predict_{box} &= \text{conv}(Conv_{box2}(Conv_{box1}(F_{box})))
\end{align}

\section{LABEL ASSIGNMENT}

We employ two different label assignment mechanisms in this study to match the model's predictions with the ground truth labels from the data: TAL \cite{tood} and SimOTA \cite{yolox} - a simpler version of OTA \cite{ota}. Both mechanisms rely on a similarity measurement function between $d_{predict}$ and $d_{truth}$ to perform the matching between them.

\subsection{Introduction}
We introduce some notations to make the presentation of the formulas below easier to follow:
\begin{itemize}
	\item $box = [l, t, r, b]$: Four pieces of information representing a bounding box.
	\item $cls = [p_1, p_2 ... p_{nclass}]$: For each $p_i$, it represents the probability of the presence of class $i$.
	\item $d_i = [box_i, cls_i]$: A data pair that includes information about the bounding box and the probabilities of labels corresponding to that bounding box.
	\item $\mathcal{T}$: The set of all $d_i$ that represents the ground truth.
	\item $\mathcal{A}$: The set off all $d_i$ that represents the model's prediction.
	\item $\mathcal{P}$: The set of all $d_i$ that represents the model's predictions matched with a $d_i$ in $\mathcal{T}$.
	\item $\mathcal{N}$: The set of all $d_i$ that represents the model's predictions not matched with any $d_i$ in $\mathcal{T}$.
	\item $\mathcal{M}(d_i)$: The set of all $d_j$ matched with $d_i$.
	\item $BCE$ : binary cross entropy function.
	\item $CIoU$ : CIoU loss fucntion \cite{ciou}.
\end{itemize}
\subsection{TAL}
The similarity measurement function between the prediction $d_{pred} \in \mathcal{A}$ and the ground truth $d_{truth} \in \mathcal{T}$ is defined as follows:

\begin{align}
	metric &= cls\_err ^ \alpha \space box\_err ^\beta\\
	cls\_err &= BCE{(cls_{pred}, cls_{truth})} \\
	box\_err &= CIoU{(box_{pred}, box_{truth})}\\
\end{align}

For each $d_{truth} \in \mathcal{T}$, match them with the $top\_k$ $d_{pred}$ having the highest $metric$, but each $d_{pred}$ can only be matched with at most one $d_{truth}$. If $d_{pred}$ is in the $top\_k$ of multiple $d_{truth}$, match $d_{pred}$ with $d_{truth}$ that has the highest $CIoU(box_{pred}, box_{truth})$. Additionally, we only consider $d_{pred}$ with the center of the receptive field located within the $box$ of $d_{truth}$ and the distance from the center of the receptive field to the center of the $box$ of $d_{truth}$ is not more than $radius$.

Typically, if a $d_{pred}$ matches with $d_{truth}$, we consider the target of $d_{pred}$ to be $d_{truth}$. The probability of the target for the corresponding classes is set to 1 if they appear, and 0 if they do not. However, in order to incorporate the idea of Qualified Loss \cite{generlized_focal_loss}, the target probabilities need to be adjusted. For classes that do not appear, the target probability is set to 0. However, for classes that do appear:

\begin{align}
	p_i &= \frac{metric(d_{pred}, d_{truth}) \mathcal{B}(d_{truth})}{\mathcal{C}(d_{truth})} \\
	\mathcal{B}(d_{i}) &= \max_{d_j \in \mathcal{M}(d_{i})}{CIoU(d_j, d_{i})} \\
	\mathcal{C}(d_{i}) &= \max_{d_j \in \mathcal{M}(d_{i})}{metric(box_j, box_{i})} \\
\end{align}

By doing this, the values of $p_i$ will be in the range $[0...1]$. $p_i$ will be greater if the $metric(d_{pred}, d_{truth})$ is larger.

\subsection{SimOTA}
\label{SimOTA-match}
Similar to TAL, SimOTA also utilizes a similarity measurement function for matching between $d_{pred}$ and $d_{truth}$:

\begin{align}
	metric &= BCE(\lambda cls_{pred}, cls_{truth}) - \alpha \log(\lambda) \\
	\lambda &= CIoU(box_{pred}, box_{truth})
\end{align}

Please note that $\alpha$ in (2) and (1) is not the same. Differing slightly from TAL, we only match $d_{truth}$ with the $top\_k$ $d_{pred}$ instances having the \textbf{smallest} $metric$. Additionally, in \cite{yolox}, SimOTA does not fix this $top\_k$ value but instead utilizes a method to estimate $top\_k$ for each $d_{truth}$. However, our experiments have shown that using dynamic $top\_k$ slows down the training process by incurring additional computational costs without bringing significant benefits. The remaining steps are also similar to those in TAL.

We also need to scale the $p_i$ values as mentioned in TAL:

\begin{equation}
	p_i = CIOU(d_{pred}, d_{truth})
\end{equation}

\section{LOSS FUNCTION}

We use two loss functions corresponding to two label assignment mechanisms: TAL and SimOTA. We will also name these two loss functions accordingly for easy tracking.

The overall loss function generally consists of two components as follows:

\begin{equation}
	\mathcal{L} = \mathcal{L}_{box} + \mathcal{L}_{cls} \
\end{equation}

Here, $\mathcal{L}_{box}$ is the loss for bounding box regression, and $\mathcal{L}_{cls}$ is the loss for label classification. There can be multiple subcomponents within these two loss functions. For those $d_{pred} \in \mathcal{N}$, they will have $\mathcal{L}_{box} = 0$, and their $cls_{truth}$ will also be completely equal to $0$.

\subsection{TAL}

TAL loss function :

\begin{equation}
	\mathcal{L} = \frac{\mathcal{L}_{box} + \mathcal{L}_{cls}}{\omega} \\
\end{equation}

Where : 

\begin{align}
	\mathcal{L}_{box} &= \delta(d_{truth})(\alpha CIoU(d_{pred}, d_{truth}) + \beta \mathcal{L}_{distribution}) \\
	\mathcal{L}_{cls} &= \gamma BCE(cls_{pred}, cls_{truth})
\end{align}

The symbol $\mathcal{L}_{distribution}$ represents the Distribution Loss function \cite{generlized_focal_loss}. And:

\begin{align}
	\delta(d_i) &= \sum_{p_j \in cls_i}{p_j} \\
	\omega &= \sum_{d_{pred} \in \mathcal{A}} \sum_{d_i \in \mathcal{M}(d_{pred})}{\delta(d_i)}
\end{align}

The symbols $\alpha, \beta, \gamma$ are hyperparameters used to scale the components of the $\mathcal{L}$ function. We fix $\alpha=7.5, \beta=1.5, \gamma=0.5$ for the TAL loss function in our experiments.

\subsection{SimOTA}
\label{SimOTA-loss}
SimOTA loss function : 

\begin{equation}
	\mathcal{L} = \frac{\mathcal{L}_{box} + \mathcal{L}_{cls}}{|\mathcal{P}|}
\end{equation}

Where :

\begin{align}
	\mathcal{L}_{box} &= \alpha CIoU(d_{pred}, d_{truth}) + \beta \mathcal{L}_{distribution} \\
	\mathcal{L}_{cls} &= exp(cls_t)|cls_{truth} - cls_{pred}|^ \nu BCE(cls_{pred}, cls_{truth})
	\label{exp(cls_t)}
\end{align}

$\mathcal{L}_{cls}$ in this case is the generalized focal loss function \cite{generlized_focal_loss}, multiplied by a class balancing factor $exp(cls_t)$. And:

\begin{equation}
	p_i \in cls_t = \begin{cases}
		class\_ratio ,& p_{truth} \neq 0 \\
		1 - class\_ratio ,& p_{truth} = 0
	\end{cases}
	\label{class_ratio}
\end{equation}

Where $class\_ratio$ is the class balancing factor defined beforehand such that classes have lower frequencies, the $class\_ratio$ will be higher. Similarly the TAL loss,  We fix $\alpha=5.5, \beta=0.5, \gamma=0.5$ and $\nu=0.5$ for the SimOTA loss function in our experiments.

\begin{table*}[!htbp]
	\centering
	\caption{Affect of balance factor and soft label technique, (mAP, GLOPs, param)}
	\label{wide-range}
	\resizebox{\textwidth}{!}{\begin{tabular}{l|c|c|c|c|c}
			\toprule
			\diagbox[width=10em, height=3.5em]{\textit{\textbf{Backbone3D}}}{\textbf{\textit{Backbone2D}}} & \textbf{YOLOv8:m} & \textbf{YOLOv8:m} & \textbf{YOLOv8:m} & \textbf{YOLOv8:m} & \textbf{YOLOv8:m} \\
			\toprule
			\multicolumn{6}{c}{\textbf{UCF101-24}}\\
			\midrule
			\textbf{ShuffleNetv2:2.0x} & 82.76 / 2.79 / 10.15 & 85.37 / 5.50 / 22.73 & 86.55 / 10.85 / 44.30 & 85.49 / 19.20 / 71.07 & 85.76 / 24.36 / 93.94  \\
			\midrule
			\textbf{ResNet:101} & 87.73 / 63.62 / 111.29 & 88.04 / 64.96 / 120.11 & 88.52 / 67.94 / 134.35 & 88.09 / 72.94 / 150.91 & 88.10 / 78.10 / 173.77    \\
			\midrule
			\textbf{ResNeXt:101} & 87.94 / 46.62 / 73.56 & 87.94 / 47.96 / 82.39 & 88.54 / 50.94 / 96.63 & 88.41 / 55.94 / 113.18 & 89.16 / 61.10 / 136.05     \\
			\midrule
			\textbf{I3D} & 88.00 / 35.52 / 36.76 & 88.22 / 36.86 / 45.58 & 88.33 / 39.84 / 59.82 & 88.34 / 44.84 / 76.38 & 88.62 / 50.00 / 99.24 \\
			\bottomrule
			\multicolumn{6}{c}{\textbf{AVAv2.2}}\\
			\midrule
			\textbf{ShuffleNetv2:2.0x} & 15.06 / 2.80 / 10.16 & 17.42 / 5.51 /22.75 & 18.29 / 10.86 / 44.33 & 18.25 / 19.21 / 71.12 & 18.35 / 24.37 / 93.98\\
			\midrule
			\textbf{ResNet:101} & 18.88 / 63.64 / 111.33 & 18.92 / 64.98 / 120.15 & 19.45 / 67.96 / 134.40 & 18.89 / 72.96 / 150.95 & 19.40 / 78.12 / 173.81 \\
			\midrule
			\textbf{ResNeXt:101} & 20.22 / 46.63 / 73.61 & 20.06 / 47.97 / 82.43 & 20.80 / 50.95 / 96.67 & 20.20 / 55.95 / 113.23 & 19.72 / 61.11 / 136.09 \\
			\midrule
			\textbf{I3D} & 17.98 / 63.64 / 111.33 & 19.09 / 36.87 / 45.62 & 20.31 / 39.85 / 59.86 & 19.42 / 44.85 / 76.42 & 19.79 / 50.01 / 99.28\\
			\bottomrule
			
	\end{tabular}}
\end{table*}
\section{EXPERIMENT SETTING}

\subsection{DATASETS}

\subsubsection{UCF101-24} Specifically designed for Human Activity Recognition (STAD) in sports. It comprises a total of 24 classes and consists of 338K labeled key frames for training purposes, along with an additional 117K key frames reserved for testing. Following the approach of YOWO and YOWOv2, we train and evaluate the model on the first split.

\subsubsection{AVAv2.2} Constructed for the challenging problem of Human Activity Recognition (STAD) and contains 184K labeled key frames for the training set and 50K key frames for the validation set. However, the test set is only used for private evaluation in the THUMOS challenge. Following YOWO and YOWOv2, we evaluate the model on the validation set.

\subsection{Implementation details}

During our experimentation, we maintained certain hyperparameter values unchanged to ensure consistency and evaluate the effectiveness of other modifications. We utilized a learning rate of 0.001 and applied linear warmup from 0 for 500 steps. Additionally, we set the weight decay to 0.0005 and set decay learning rate factor of 2. The batch size was set to 8, and we accumulated gradients over 16 iterations. The input clips had a length of 16 frames, with each frame resized to 224x224 pixels. Based on these settings, we evaluated the model's performance using mAP (mean Average Precision) and GLOPs (Giga Operations per Second) as metrics. Furthermore, we measured the model's prediction speed in frames per second (FPS).

For UCF101-24, TAL was utilized for label assignment and the loss function. The model underwent training for 7 epochs, with the learning rate decay occurring post the 1st, 2nd, 3rd, 4th, and 5th epochs.

For AVAv2.2, SimOTA was employed for label assignment and the loss function. The model was trained for 9 epochs, with the learning rate decay applied after the 3rd, 4th, 5th, and 6th epochs.

We experimented with various configurations using the YOWOv3 framework; however, we only selected three most prominent models, namely YOWOv3-Tiny (YOLOv8 nano, shufflenetv2), YOWOv3-Medium (YOLOv8 medium, shufflenetv2), and YOWOv3-Large (YOLOv8 medium, I3D) as the three primary model in this study..

\subsection{Metric}

We use the metric mean Average Precision with an IoU threshold of 0.5. On AVAv2.2, we only evaluate on a subset of 60 action classes, following the official procedure on Task B – Spatio-temporal Action Localization (AVA) of Activity challenge \cite{ActivityNet}.

\section{PERFORMANCE ANALYSIS}
\subsection{Compare with YOWO and YOWOv2}

We compare our YOWOv3 model with its predecessors, YOWOv2 and YOWO. Figure \ref{trade-off} visually shows how computational resources trade off with performance on the UCF101-24 dataset, while Table \ref{compare} summarizes results for both UCF101-24 and AVAv2.2. Looking at Table \ref{compare}, despite YOWOv2 improving mAP, it doesn't efficiently use computational resources, as GLOPs increase by about 22\% compared to YOWO (in the L - Large model). On the other hand, with YOWOv3, especially YOWOv3-L, we achieve comparable performance to YOWOv2-L while reducing GLOPs by around 26\% and needing only 54.5\% of the parameters. This demonstrates the effectiveness of the proposed YOWOv3 framework.
\begin{table*}[!hbpt]
	\centering
	\caption{Affect of dynamic and fixed top\_k, mAP(\%)}
	\label{EMA}
	\begin{tabular}{c|c|c|c|c|c|c}
		\toprule
		& \multicolumn{3}{c|}{\textbf{with EMA}} & \multicolumn{3}{c}{\textbf{without EMA}} \\
		\midrule
		\diagbox[width=5em, height=3.5em]{\textit{\textbf{Epoch}}}{\textbf{\textit{Model}}} & \textbf{YOWOv3-T} & \textbf{YOWOv3-M} & \textbf{YOWOv3-L} & \textbf{YOWOv3-T} & \textbf{YOWOv3-M} & \textbf{YOWOv3-L} \\
		\toprule
		\textbf{1} & 78.57 & 83.45 & 86.23 & 77.93 & 82.47 & 85.9   \\
		\textbf{2} & 81.93 & 85.3  & 88.15 & 81.25 & 84.32 & 87.69  \\
		\textbf{3} & 82.43 & 86.28 & 88.16 & 81.26 & 85.76 & 87.38  \\
		\textbf{4} & 82.41 & 86.67 & 88.37 & 81.81 & 86.19 & 88.35  \\
		\textbf{5} & 82.65 & 86.51 & 88.28 & 81.87 & 86.29 & 87.9   \\
		\textbf{6} & 82.85 & 86.53 & 88.37 & 82.87 & 86.33 & 88.35  \\
		\textbf{7} & 82.76 & 86.55 & 88.33 & 82.58 & 86.39 & 88.26  \\
	\end{tabular}
\end{table*}
\begin{table}[!h]
	\centering
	\caption{Compare the performance of models on the UCF101-24 and AVAv2.2 datasets}
	\label{compare}
	\begin{tabular}{l|c|c|c|c}
		\toprule
		\textbf{Model}    & \textbf{UCF24}  & \textbf{AVA}    & \textbf{GLOPs} & \textbf{Params(M)} \\
		\midrule
		YOWO     & 80.4   & 17.9   & 43.7  & 121.4     \\
		YOWOv2-T & 80.5   & 14.9   & 2.9   & 10.9      \\
		YOWOv2-M & 83.1   & 18.4   & 12    & 52        \\
		YOWOv2-L & 85.2   & 20.2   & 53.6  & 109.7     \\
		\midrule
		YOWOv3-T & 82.76  & 15.06  & 2.8   & 10.1      \\
		YOWOv3-M & 86.55  & 18.29  & 10.8  & 44.3      \\
		YOWOv3-L & \textbf{88.33} & \textbf{20.31} & \textbf{39.8} & \textbf{59.8} \\
		\bottomrule
	\end{tabular}
\end{table}

\subsection{Ablation study}

\subsubsection{Effectiveness of class balance loss}
AVAv2.2 is an extremely imbalanced dataset. For instance, each of the 3 classes "listen to," "talk to," and "watch" appear over 100K times in the entire training set, while classes like "fight/hit," "give/serve," or "grab" have fewer than 3K occurrences each. This requires additional techniques to mitigate the impact of this imbalance if we aim to improve the overall mAP.

We employed two techniques to address this issue: soft labels (Qualified Loss) as introduced in section 4.2 and the inclusion of a class balance term as discussed in section 5.2. The effects of these two components are summarized in Table \ref{soft-balance}.

\begin{table}[h]
	\centering
	\caption{Affect of balance factor and soft label technique (mAP \%)}
	\label{soft-balance}
	\begin{tabular}{l|c|c|c}
		\toprule
		\textbf{Model}    & \textbf{Both}  & \textbf{w/o balance}    & \textbf{w/o soft label} \\
		\midrule
		YOWOv3-T & 15.06   & 13.82   & \textbf{15.16}     \\
		YOWOv3-M & \textbf{18.29}   & 16.8    & 17.7      \\
		YOWOv3-L & \textbf{20.31}   & 18.43   & 20.07     \\
		\bottomrule
	\end{tabular}
\end{table}

For the class balance term, specifically the $class\_ratio$ in equation \eqref{class_ratio}, we set this value such that for classes appearing more frequently, it will be closer to $0.5$, while for less common classes, it will closer to $0$. In other words, the range of values for $class\_ratio$ will approximately fall within the range of $(0 ... 0.5)$, with denser classes tending towards $0.5$ and less frequent classes tending towards $0$. In this way, $exp(cls\_t)$ in equation \eqref{exp(cls_t)} behaves as follows: for classes with a $class\_ratio$ close to 0.5 (meaning they appear more frequently), the class balance term $exp(cls\_t)$ will remain almost unchanged, meaning that the loss will not change significantly if mispredicted. Conversely, for classes with a $class\_ratio$ close to $0$ (indicating they are less common), the class balance term $exp(cls\_t)$ will heavily penalize if a class present in the ground truth is not predicted by the model, while the penalty will be much lighter if a class not in the ground truth is predicted. This creates a bias that helps improve the prediction of less common classes. Additionally, soft labels are applied in the hopes of reducing the impact of the model's overconfidence, particularly towards classes that appear too frequently.

Table \ref{soft-balance} demonstrates that the class balance term plays a crucial role and has the most significant impact on the overall results, whereas soft labels have a weaker effect and prove to be effective with more complex models.

\subsubsection{Effectiveness of dynamic top k}
\begin{table}[!htbp]
	\centering
	\caption{Affect of dynamic and fixed top\_k(mAP \%)}
	\label{dynamic}
	\resizebox{0.45\textwidth}{!}{\begin{tabular}{l|c|c|c|c}
			\toprule
			\textbf{Model}    & \textbf{dynamic}  & \textbf{k=5}    & \textbf{k=10} & \textbf{k=20} \\
			\midrule
			YOWOv3-T & \textbf{15.18}   & 14.61   & 15.06   & 14.08      \\
			YOWOv3-M & 17.43   & 16.79   & \textbf{18.29}   & 17.74        \\
			YOWOv3-L & 19.79   & 19.19   & \textbf{20.31}   & 19.62     \\
			\bottomrule
	\end{tabular}}
\end{table}
For label assignment one to many, a ground truth box will be matched with multiple predicted boxes. The number of boxes chosen for matching can either be dynamically estimated or predetermined. The impact of this approach has been experimented with and summarized in Table \ref{dynamic}.

The experiments show that selecting a fixed $k$ yields slightly better results. Furthermore, automatically estimating the $top\_k$ results in an additional $6\%$ increase in training time. These results encourage treating the $top\_k$ as a hyperparameter rather than opting for an estimation method.
\subsubsection{Effectiveness of EMA}

 In addition to the model's base weights, we preserved an Exponential Moving Average (EMA) variant of the model weights and evaluated the EMA's impact, as detailed in Table \ref{EMA}. The results show that EMA significantly influences the model's performance in the initial epochs and has less impact in later epochs. This indicates that EMA helps the model converge faster in the early stages and slightly improves the mAP score in the later stages of the training process.
\begin{figure*}[!htbp]
	\begin{minipage}{\textwidth}
		\centering
		\begin{subfigure}{0.23\textwidth}
			\centering
			\includegraphics[width=\linewidth]{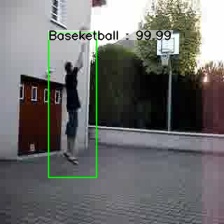}
			%\caption{Caption for Image 1}
			%\label{fig:img1}
		\end{subfigure}
		\hfill
		\begin{subfigure}{0.23\textwidth}
			\centering
			\includegraphics[width=\linewidth]{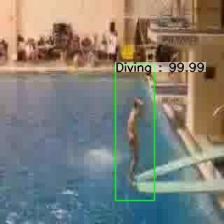}
			%\caption{Caption for Image 2}
			%\label{fig:img2}
		\end{subfigure}
		\hfill
		\begin{subfigure}{0.23\textwidth}
			\centering
			\includegraphics[width=\linewidth]{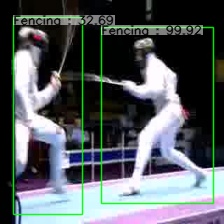}
			%\caption{Caption for Image 3}
			%\label{fig:img3}
		\end{subfigure}
		\hfill
		\begin{subfigure}{0.23\textwidth}
			\centering
			\includegraphics[width=\linewidth]{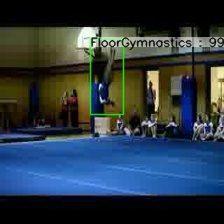}
			%\caption{Caption for Image 4}
			%\label{fig:img4}
		\end{subfigure}
	\end{minipage}
	
	\begin{minipage}{\textwidth}
		\centering
		\begin{subfigure}{0.23\textwidth}
			\centering
			\includegraphics[width=\linewidth]{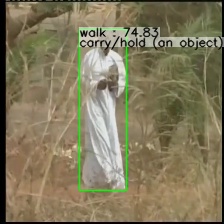}
			%\caption{Caption for Image 5}
			%\label{fig:img5}
		\end{subfigure}
		\hfill
		\begin{subfigure}{0.23\textwidth}
			\centering
			\includegraphics[width=\linewidth]{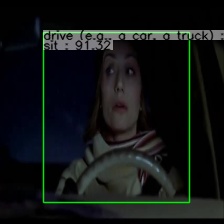}
			%\caption{Caption for Image 6}
			%\label{fig:img6}
		\end{subfigure}
		\hfill
		\begin{subfigure}{0.23\textwidth}
			\centering
			\includegraphics[width=\linewidth]{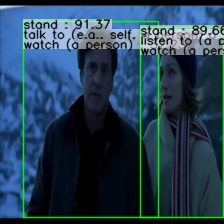}
			%\caption{Caption for Image 7}
			%\label{fig:img7}
		\end{subfigure}
		\hfill
		\begin{subfigure}{0.23\textwidth}
			\centering
			\includegraphics[width=\linewidth]{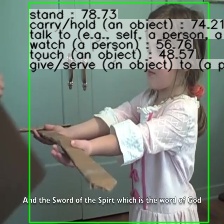}
			%\caption{Caption for Image 8}
			%\label{fig:img8}
		\end{subfigure}
	\end{minipage}
	
	\caption{Visualization of YOWOv3 on UCF101-24 and AVAv2.2}
	\label{fig:overall}
\end{figure*}
\subsection{Wide range experiments}

We examined several designs and configurations using the UCF101-24 and AVAv2.2 datasets in our study of the YOWOv3 framework. Through this process, a wide range of pretrained resources, from simple models to more sophisticated ones that can handle challenging tasks, were developed to meet the needs of the community. The findings, which are outlined in Table 1, demonstrate our dedication to sharing knowledge.  These essential pretrained resources are now publicly available, reflecting our dedication to advancing the field of AI research. Each cell in the table includes three pieces of information: mAP, GLOPs, and parameters, respectively.

\subsection{Visualization}

We provide several visualizations to offer a clear insight into the model's performance. The examples are taken from the evaluation sets of two datasets, UCF101-24 (top row) and AVAv2.2 (bottom row). These visualizations are presented in Figure \ref{fig:overall}.

\section{Conclusion}

In conclusion, our paper introduces a new framework called YOWOv3 aimed at providing the research community with an efficient framework for the Spatial-Temporal Action Detection problem, meeting the requirements for limited computational capabilities in practical applications. Drawing on the concept of the Two-Stream Network, we have effectively combined network architectures and enhancement techniques to achieve the research objectives. On the UCF101-24 dataset, YOWOv3 achieved an mAP score of 88.33\%, and on the AVAv2.2 dataset, it reached 20.31\%, surpassing previous models using the same approach while reducing the number of parameters by 45.5\% and GLOP by 25.74\%. With this newly proposed framework, we hope to drive future research and practical applications, delivering value to the community.

\begin{acknowledgments}
	This research was supported by the scientific research fund of
	University of Information Technology (UIT-VNUHCM).
\end{acknowledgments}

%%
%% Define the bibliography file to be used

%\bibliography{sample-ceur.bib}
%\begin{thebibliography}
%\end{thebibliography}
%%
%\newpage
%\bibliography{sample-ceur}

\end{document}